\documentclass[letterpaper, 10 pt, conference]{ieeeconf}  

\IEEEoverridecommandlockouts                              

\usepackage{lipsum} 
\usepackage{epsfig}
\usepackage{float}
\usepackage{bbold}
\usepackage{bm}
\usepackage{hyperref}

\usepackage[noadjust]{cite}

\usepackage{color,soul} 

\usepackage{amsmath}

\title{\LARGE \bf
Speeding up 6-DoF Grasp Sampling with Quality-Diversity
}

\begin{document}

\author{Johann Huber$^{*1}$, François Hélénon$^{*1}$, Mathilde Kappel$^{*1}$, \\
Elie Chelly$^{1}$, Mahdi Khoramshahi$^{1}$, Faïz Ben Amar$^{1}$ and Stéphane Doncieux$^{1}$ 
\thanks{$^*$ equal contribution and corresponding authors}
\thanks{$^{1}$Sorbonne Université, CNRS, Institut des Systèmes Intelligents et de Robotique, ISIR, F-75005 Paris, France {\tt\small \{helenon, huber, kappel, chelly, khoramshahi, benamar, doncieux\}@isir.upmc.fr}}}

\maketitle
\thispagestyle{empty}
\pagestyle{empty}





\begin{abstract}

Recent advances in AI have led to significant results in robotic learning, including natural language-conditioned planning and efficient optimization of controllers using generative models. However, the interaction data remains the bottleneck for generalization. Getting data for grasping is a critical challenge, as this skill is required to complete many manipulation tasks. Quality-Diversity (QD) algorithms optimize a set of solutions to get diverse, high-performing solutions to a given problem. This paper investigates how QD can be combined with priors to speed up the generation of diverse grasps poses in simulation compared to standard 6-DoF grasp sampling schemes. Experiments conducted on 4 grippers with 2-to-5 fingers on standard objects show that QD outperforms commonly used methods by a large margin. Further experiments show that QD optimization automatically finds some efficient priors that are usually hard coded. The deployment of generated grasps on a 2-finger gripper and an Allegro hand shows that the diversity produced maintains sim-to-real transferability. We believe these results to be a significant step toward the generation of large datasets that can lead to robust and generalizing robotic grasping policies.

\end{abstract}





\section{INTRODUCTION}

Grasping is a skill of great interest in robotics as it is a prerequisite for many manipulation tasks \cite{hodson2018gripping}. The previously prevalent analytical-based methods \cite{nguyen1988constructing} are gradually giving way to data-driven strategies since the beginning of the 21st century \cite{zhang2022robotic}. However, the challenging exploration aspect of grasping hinders the bootstrapping of the learning process, as most grasps attempted by a policy initialized at random yield no reward \cite{huber2023quality}. Many studies have addressed this issue by employing imitation learning \cite{qin2022from,wang2021demograsp,sefat2022singledemograsp}, parallel grippers \cite{depierre2018jacquard,fang2020graspnet}, and top-down movements \cite{levine2018handeye,mahler2017dexnet2,yang2023pave}. However, these approaches restrict the operational space, constraining the policies' adaptability.

The recent advances in robotic learning demonstrate the capabilities of data-based approaches for skill acquisition \cite{chi2023diffusion,octo2023octo,urain2023se3diffusionfield,barad2023graspldm,chen2024nsgf}. These results rely on modern artificial intelligence methods that require a tremendous amount of high-quality data to generalize to unknown scenes. It has led to the release of large datasets \cite{padalkar2023openxembodiement}, many of them focusing on grasping \cite{fang2020graspnet,eppner2021acronym,turpin2023fastgraspd}. Acquiring such high-quality datasets is becoming critical to allow high representational power architecture to achieve generalization \cite{chi2023diffusion} or to build efficient foundation models \cite{octo2023octo} to transfer skills between platforms and scenarios.

Great efforts have been made to build large sets of real data \cite{fang2020graspnet,padalkar2023openxembodiement}, but the acquisition of such data is very slow and expensive. Recent works leverage simulated scenes to speed up the data generation process \cite{eppner2021acronym,turpin2023fastgraspd,huber2023quality} while assuring the sim-to-real transfer through dedicated quality criteria \cite{huber2023domainrandomization}. 

The advent of data-driven approaches has led to the formalization of grasping as a 6 Degrees-of-Freedoms (6-DoF) pose synthesis problem that can be predicted by a trained model \cite{newbury2023dlgraspsynthesissurvey}. Recently, Eppner et al. \cite{eppner2023abw2g} conducted a systematic study to identify the best grasp sampling schemes among the most commonly used ones in the literature. In the Evolutionary Algorithms field, Huber et al. \cite{huber2023quality} studied how Quality-Diversity (QD) algorithms can be applied to generate diverse datasets of robust reach-and-grasp trajectories. 

This paper shows that QD methods can significantly speed up the 6-DoF grasp sampling. In particular:

\begin{figure}[t]
  \centering
\centering
  \includegraphics[width=0.8\columnwidth]{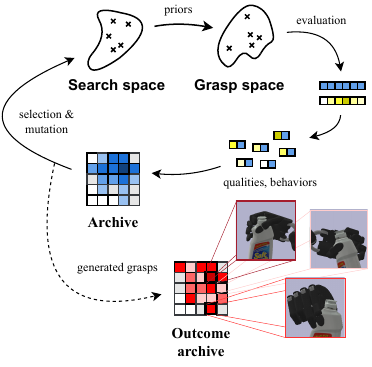}
  \caption{\textbf{Overview of the proposed framework.} It consists of a population-based algorithm that allows the efficient exploration of the space of possible grasp positions from a genotype space. The optimization process is driven by searching for diverse and high-performing grasps using a structured archive of previously found solutions \cite{cully2022qd}. The genotype space is designed to leverage the robotic priors that are commonly used in 6-DoF grasp sampling \cite{eppner2023abw2g}. All grasps ever produced are added to an outcome archive, which is the output of the algorithm.}
  \label{fig:qdg6dof_overview}
\end{figure}

\begin{itemize}
    \item We propose a framework that combines robotic priors with QD algorithms to generate large sets of diverse and robust grasps;
    \item We show that variants based on this framework outperform state-of-the-art 6-DoF grasp synthesis schemes on 4 grippers and a dozen of standard objects in simulation;
    \item Experiments conducted on a physical Franka Emika Panda gripper and an Allegro hand show that the proposed method generate grasps that successfully transfer into the real world while preserving diversity. 
\end{itemize}

The code had been made publicly available\footnote{\url{https://gitlab.isir.upmc.fr/l2g/qd\_grasp\_6dof}}. More details can be found on the project website\footnote{\url{https://qdgrasp.github.io/}}.


\section{RELATED WORKS}
\label{sec:2_related_works}



\textbf{\textit{Learning to Grasp in Robotics.}} Many paradigms have been explored since the advent of data-driven approaches in the field, including reinforcement learning \cite{chen2023rlgrasp1,zhou2023rlgrasp3}, and learning from a few demonstrations \cite{wang2021demograsp,sefat2022singledemograsp}. However, these methods require constraining the search space to make the problem tractable. Many data-greedy approaches were proposed to get generalization capabilities \cite{levine2018handeye,mahler2017dexnet2,fang2020graspnet}. Recently were leveraged Diffusion models \cite{urain2023se3diffusionfield}, Variational autoencoders \cite{barad2023graspldm}, or 
neural representations \cite{chen2024nsgf} to sample grasps poses on unknown objects. Interestingly, most of them exploit automatically generated data in their learning process. 


\textbf{\textit{Automatic Generation of Grasping Datasets.}} Datasets for robotic learning can be directly collected in the real world \cite{levine2018handeye,padalkar2023openxembodiement}, but this process is very time and cost-expensive. To circumvent this problem, many works focus on simulation. Data were originally annotated with analytic criteria \cite{godlfeder2009columbia,mahler2017dexnet2}, which are still used nowadays for top-down grasps \cite{vuong2023graspanything}. On the contrary, more and more works rely on a physics engine to simulate the gripper-object interaction. These methods leverage robotic priors to speed up the data generation, resulting in separated studies for 2-finger \cite{depierre2018jacquard,eppner2021acronym,eppner2023abw2g} and multi-fingered \cite{turpin2023fastgraspd} grippers. 

However, there is no consensus on the best methods to generate diverse grasps for n-finger grippers efficiently. Eppner et al. \cite{eppner2023abw2g} put a stepstone on that matter through a systematic comparison of standard sampling schemes for 6-DoF grasps poses on a parallel gripper. We extend this analysis to n-finger by including new QD-based sampling schemes.



\textbf{\textit{Quality Diversity.}} 
Quality-Diversity methods are optimization algorithms that aim to generate a set of diverse and high-performing solutions to a given problem \cite{cully2022qd}. Recently, Huber et al. \cite{huber2023quality} demonstrated that QD could generate datasets of diverse and high-performing reach-and-grasp trajectories for n-finger grippers and proposed a Domain-Randomization-based method for estimating the probability of successfully completing the sim-to-real transfer \cite{huber2023domainrandomization}. We extend this work to 6DoF grasp sampling to investigate how QD can speed up the automatic generation of grasping data.

\begin{figure}[t]
  \centering
  \includegraphics[width=0.8\columnwidth]{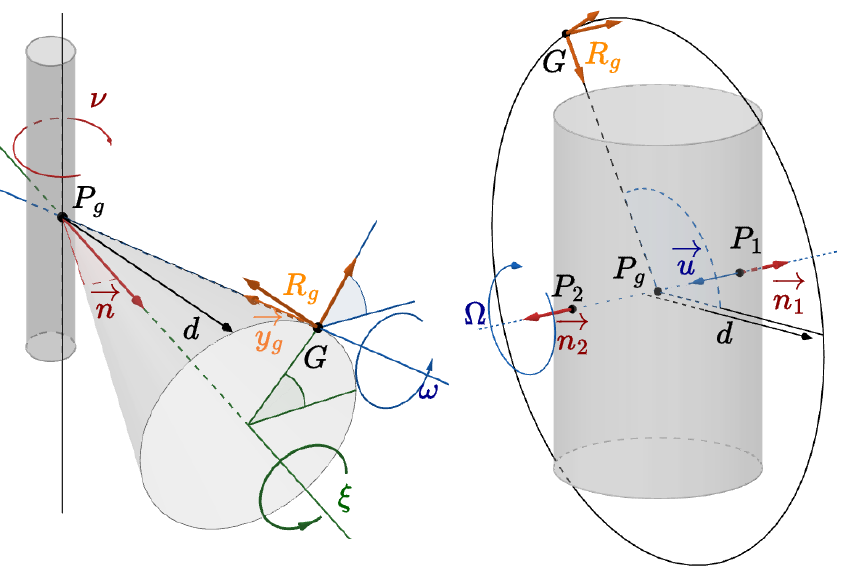}
  \caption{\textbf{Commonly used priors in 6DoF grasp sampling.} (Left) approach-based ; (Right) antipodal-based. $R_g$ is the frame associated with the gripper. The cylinder is an illustration of a targeted object.}
  \label{fig:notations_approach_antipodal}
\end{figure}


\section{METHOD}


\subsection{6DoF Grasp Sampling}
\label{sec:III_A_6dof_grasp_sampling}

Let $\Theta$ be the \textit{parameter space}, and $\theta \in \Theta$ an \textit{individual}. Let $G$ be the space of possible grasps given a rigid object and a gripper, and $\mathcal{G}\subseteq G$ be the space of successful grasps. This section aims to define how robotic priors can be used to project elements from $\Theta$ to $G$.


\textbf{\textit{Definition.}} Let $g\in SE(3) \times \mathbb{R}^n$ be a grasp, with $n$ being the number of internal DoF. To reduce the search space on multi-fingered grippers, we close each hand with synergies. However, some grippers need flexibility in the initial joint pose to exploit their gripping capabilities. Thus, we have $g\in SE(3) \times S_y \times \mathbb{R}^k$ with $S_y\subseteq \mathbb{N}^{+*}$ the space of predefined synergies with $\text{Card}(S_y)=m$ the number of synergies for a given hand, and $k$ the number of joints for which the initial state is included in the search space. For a parallel gripper, we have $m=1$ and $k=0$, and thus $g\in SE(3)$, similarly to \cite{eppner2023abw2g}. Details on the synergies and controlled initial joints are provided in the supplementary materials.


\textbf{\textit{Prior-based Sampling schemes.}} The present article relies on the most commonly used schemes for 2-finger grippers \cite{eppner2023abw2g}. To our knowledge, the commonly used methods to generate diverse multi-fingered grasps rely on several optimization steps \cite{yao2023allegrograsp,turpin2022graspd,turpin2023fastgraspd}. The present work aims to compare methods that produce grasps poses by directly sampling in a search space without further optimization steps. We let multi-steps methods for future work. The sampling schemes evaluated by Eppner et al. include the \textit{antipodal-based} and \textit{approach-based} ones (Fig. \ref{fig:notations_approach_antipodal}). To have a meaningful random baseline, we also introduce a \textit{contact-based} sampling scheme. Let ${R_{g}}=(\vec{x}_g, \vec{y}_g, \vec{z}_g)$ be the frame associated with the end effector such that $(\vec{z}_g, \vec{x}_g)$ generates the palm plane and $\vec{y}_g$ drives the gripping direction. 

Approach-based prior: Let $\vec{n}$ be the normal on the object surface at the targeted reference point $P_g$. This method aims to align $\vec{n}$ with $\vec{y_g}$ such that the angle verifies:
\begin{equation}
    \label{eq:nu_def}
    (\vec{n}, -\vec{y}_g) = \nu \leq \nu_{a}
\end{equation}
where $\nu_{a}$ is the maximum half-aperture of the approach cone. The parameter $\theta$ to explore $G$ is thus:
\begin{equation}
    \theta = (x_r, y_r, z_r, d, \nu, \xi, \omega)
\end{equation}
where $P_g=(x_r, y_r, z_r)$ defines the targeted reference pose, $d$ is the gripper distance to contact point, $\nu$ is defined as in eq. (\ref{eq:nu_def}), $\xi$ is the cone revolution angle around $\vec{n}$, and $\omega$ is the gripper rotation around $\vec{y}_g$. The maximum half-aperture has been set to $\nu_a=\pi/4$ , like in Eppner et al. \cite{eppner2023abw2g}.

Antipodal-based prior: Let $\vec{u}=-\vec{n}$, the opposite vector to the normal $\vec{n}$ at the first contact point $P_1=(x_{r_1}, y_{r_1}, z_{r_1})$. The last point found on the object surface along $\vec{u}$ defines the second contact point $P_1$. The normal of the object surface at $P_1$ is noted $\vec{n}_1$. This method consists of finding gripper poses that apply opposite forces on $P_2$ and $P_1$. Therefore, the following criterion must be verified:
\begin{equation}
    \label{eq:nu_def}
    (\vec{n}_1, -\vec{n}_2) \leq \nu_t
\end{equation}
where $\nu_t$ is the maximal allowed difference between $\vec{n}_1$ and $\vec{n}_2$ to define an antipodal grasp. We set $\nu_t = \pi/6$. The parameter $\theta$ is thus:
\begin{equation}
    \theta = (x_{r_1}, y_{r_1}, z_{r_1}, \Omega)
\end{equation}
where $\Omega$ is the gripper orientation around $\vec{u}$.

The contact-based methods have been introduced to get baselines with minimal priors. It consists of the approach-based method with $\nu_a=\pi$. Consequently, sampling the parameter space can easily make the gripper overlap with the object. If the approach assumption is as efficient as reported in the literature \cite{eppner2023abw2g}, the contact-based variants should be significantly less sample efficient – while avoiding gripper poses that are too far from the object to grasp it.


\subsection{Leveraging Quality-Diversity for 6DoF-Grasp Sampling}

This section aims to detail how we can explore $G$ from $\Theta$ to discover solutions in $\mathcal{G}$. 


\textbf{\textit{Definition.}} This study uses QD standard notations \cite{cully2022qd}. Let $\mathcal{B} \subseteq \mathbb{R}^{n_b}$ be the \textit{behavior space}, and $\phi_{\mathcal{B}}:\Theta \rightarrow \mathcal{B}$ the \textit{behavior function}, which assigns a \textit{behavior descriptor} $b_\theta = \phi_{\mathcal{B}}(\theta)$ to each $\theta$. The \textit{fitness function} is $f:\Theta\rightarrow \mathbb{R}$, and $d_{\mathcal{B}}:\mathcal{B}^2 \rightarrow \mathbb{R}$ is a distance function within $\mathcal{B}$. The goal is to generate an \textit{archive} $A$ such that:
\begin{equation}
\left\{\begin{matrix}
\forall b \in \mathcal{B}_{reach}, \, \exists \theta \in A, \, d_{\mathcal{B}}(\phi_{\mathcal{B}}(\theta), b) < \epsilon \\
\forall \theta' \in A, \, \theta' =  \text{argmax}_{\theta\in N(b_{\theta'})}f(\theta) 
\end{matrix}\right.
\label{eq:qd_background}
\end{equation}
where $\mathcal{B}_{reach} \subseteq \mathcal{B}$ is the \textit{reachable behavior space}, $\epsilon\in\mathbb{R}^{+*}$ defines the density of $\mathcal{B}_{reach}$ paving, and $N(b_{\theta'})= \{ \theta \mid neighbor_{d_{\mathcal{B}}}(b_\theta, b_{\theta'}) \}$ is the set of solutions with close projections in $\mathcal{B}$. $\phi_{\mathcal{B}}$ is deterministic.


\textbf{\textit{Efficiently exploring $\bm{G}$ from $\bm{\mathit{\Theta}}$.}} The overall algorithmic principle is presented in Fig. \ref{fig:qdg6dof_overview}. It consists of a standard QD framework, where a mutation-selection process optimizes an archive $A$ of solutions. Individuals are projected to the grasp space by exploiting the previously described priors. Depending on the gripper, the individual $\theta$ has additional values to search among the space of possible joint initial positions and synergies.

To search within the space of contact points on the object surface, the mesh is first preprocessed to get a uniform coverage of its surface. Let $S_c$ be the resulting set of contact points for a given object. The first components of $\theta$ are in practice used to define the position $(x_{c_f}, y_{c_f}, z_{c_f})$ of a contact point finder $P_f$. The first targeted reference point (either $P_g$ or $P_1$ depending on the prior) is defined as the closest point to $P_f$ in $S_c$. This method has been used to allow a computationally efficient, physically meaningful local search. A direct search within a flattened list of contact points would limit the locality to the way the mesh has been built, such that two points far away in the list can, in practice, be very close on the object's surface.

The components of individuals are normalized to lie in $[-1,1]$ by limiting the search space to the bounding box of the targeted object. Each component is projected in its own interval when projecting $\theta$ from $\Theta$ to $G$ to eventually get a grasping $g\in G$ that consists of a 6DoF pose in $SE(3)$ and a constant closure strategy (defined by the initial joint states and the synergy).


\textbf{\textit{Evaluation.}} After a population of genomes has been sampled, the computed corresponding grasps are evaluated in simulation. A 3d model of the end effector is set to the computed position in $SE(3)$. The position is already discarded if there is an overlapping between the end effector and the object. Otherwise, a constant closure is then applied to the gripper, with respect to the individual additional components if necessary (i.e. initial joint states and synergy). If the fingers are in contact with the object, a verification phase is conducted – similarly to Eppner et al. \cite{eppner2023abw2g}. It consists of applying a fixed shaking pattern on the end effector. The quality of the grasp (called the \textit{fitness} $f$) is computed as the number of shakes the grasp $g$ has resisted.


\textbf{\textit{Population-based optimization.}} Quality-Diversity methods rely on a behavioral characterization associated with each evaluated solution \cite{cully2022qd}. We extend \cite{huber2023quality} formulation of the reach-and-grasp problem to the 6-DoF grasp pose sampling problem by exploiting the same behavior space $\mathcal{B}$, which is the Cartesian position of the end effector when applying the forces on the object surface. Each QD method has its own mutation-selection pattern. An \textit{outcome archive} $A_o$ is distinguished from the optimization archive \cite{huber2023quality}. The output $A_o$ is eventually filled with a large variety of robust grasps.


\begin{figure}[t]
  \centering
  \includegraphics[width=\columnwidth]{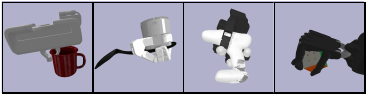}
  \caption{\textbf{Considered scenes.} Grasp sampling schemes are evaluated on YCB objects \cite{calli2015benchmarking} in the Pybullet \cite{coumans2016pybullet} simulators. Experiments involve a FE Panda gripper, a Barrett hand, an Allegro hand and a Shadow hand.}
  \label{fig:bullet_scenes}
\end{figure}

\section{EXPERIMENTS}   

This section provides the most essential information about the experiments. Details of hyperparameters and implementations are provided in Appendix \ref{sec:a1_exp_details_sim}.


\textbf{\textit{Grippers and objects.}} A comparative analysis has first been conducted in the simulator Pybullet \cite{coumans2016pybullet} (Fig. \ref{fig:bullet_scenes}). It involves 4 grippers: a Franka Emika Panda 2-finger gripper, a 3-finger Barrett hand 280, a 4-finger Allegro hand, and a 5-finger Shadow hand. The multi-fingered grippers are controlled with synergies to remain in the 6-DoF grasp pose paradigm. A subset of 10 objects from the YCB object dataset \cite{calli2015benchmarking} were used to evaluate the sampling.


\textbf{\textit{Compared methods.}} The following methods are compared: \textit{contact\_rand}, \textit{approach\_rand} and \textit{antipodal\_rand}. They are the standard prior-based sampling schemes described in section \ref{sec:III_A_6dof_grasp_sampling}. Elements are randomly sampled from the search space, similarly to Eppner et al. \cite{eppner2023abw2g}. Some variants of QD methods with robotic priors have also been compared: \textit{ME\_scs} \cite{huber2023quality}, which is state-of-the-art for reach-and-grasp trajectory generation, and \textit{CMA\_MAE} \cite{fontaine2023cmamae}, a general purpose state-of-the-art QD algorithm. \textit{ME\_rand} \cite{mouret2015mapelites} has also been added as a baseline. Note that all prior-based methods are reported with a dedicated prefix (\textit{approach\_*}, \textit{antipodal\_*}, or \textit{contact\_*}). QD methods without prefixes rely on a genome that directly encodes $g$ with spherical coordinates and Euler angles w.r.t. the object frame.  

The 3 kinds of priors are studied on the 2-finger gripper. However, the antipodal grasps as described in Eppner et al. are not applicable to multi-fingered grippers. The antipodal-based methods are thus discarded for the other hands. Note that variants of approach-based methods are applied in the literature to generate grasp poses for multi-fingered grippers \cite{roa2012saisiedlr,grimm2021saisiekit,grimm2021saisiekit2}.


\textbf{\textit{Evaluation metrics.}} To identify the best 6-DoF grasp sampling methods for multiple grippers, we compute an estimation of the successful grasp space coverage $cvg(\mathcal{G})$, similarly to Eppner et al. \cite{eppner2023abw2g}. To do so, 5 runs of 10M evaluated samples for \textit{ME\_scs} and \textit{approach\_me\_scs} have been computed. All the generated grasps are then added to a single set of successful grasps $S_{\mathcal{G}}$. To limit the size of $S_{\mathcal{G}}$, we only keep the Cartesian position of the end effector while applying the forces on the object surface and round it to $0.01 $ m. $S_{\mathcal{G}}$ contains a single occurrence of each grasp pose. The \textit{grasp pose coverage} $cvg(\mathcal{G})$ is computed by extracting the Cartesian poses of each successful grasp found in the outcome archive $A_o$. Positions are then rounded to $0.01 $ m. The coverage is calculated as the ratio of elements found in $S_{\mathcal{G}}$ divided by $\text{Card}(S_{\mathcal{G}})$. Therefore, $cvg(\mathcal{G}) \in [0,1]$, with $cvg(\mathcal{G})=1$ being the optimal value.


\section{RESULTS AND DISCUSSION}

\begin{figure}[t]
  \centering
  \includegraphics[width=\columnwidth]{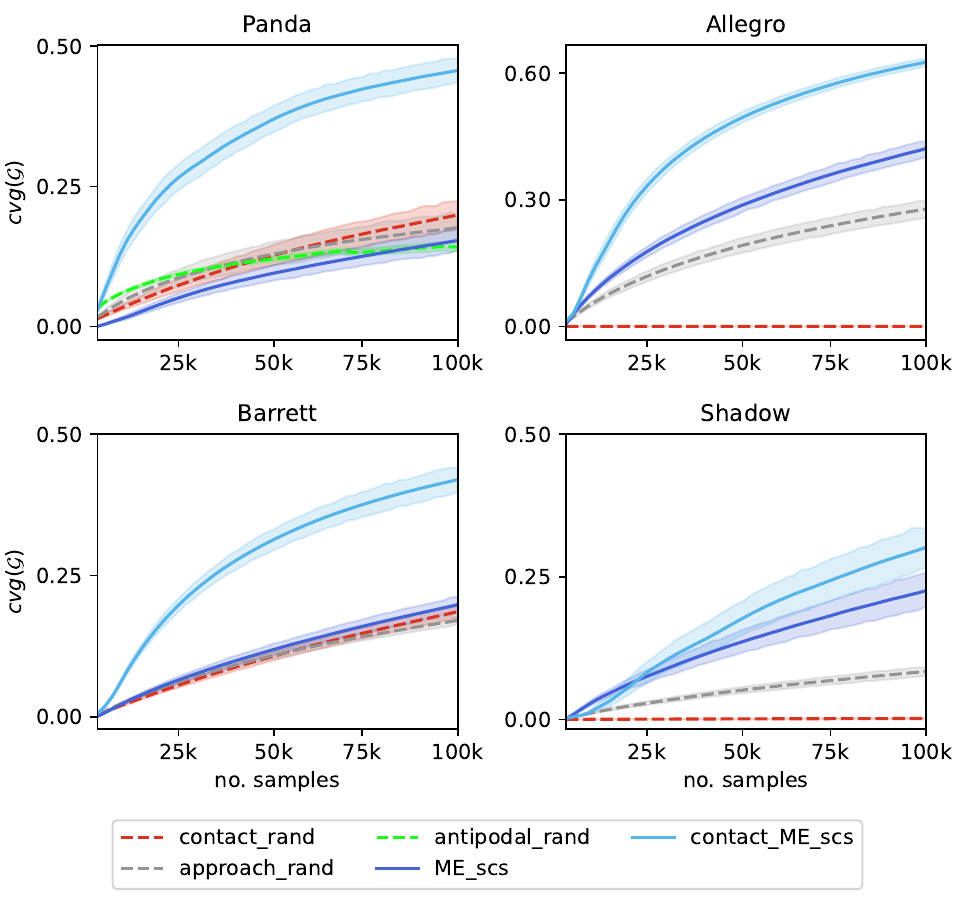}
  \caption{\textbf{QD-based vs standard prior-based sampling schemes.} Ratio of diverse, successful grasp found w.r.t. the number of evaluated samples. The dashed lines are commonly used in the literature; the plain ones are QD-based. \textit{contact\_ME\_scs} outperforms the standard schemes and the raw QD method by a large margin on the 4 considered grippers.}
  \label{fig:exp1_results}
\end{figure}

\begin{figure}[t]
  \centering
  \includegraphics[width=\columnwidth]{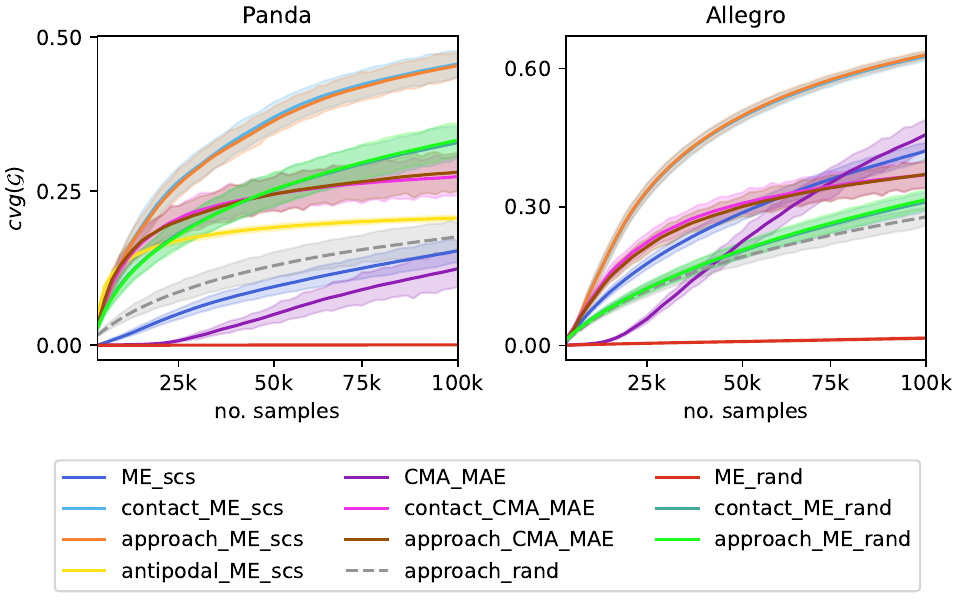}
  \caption{\textbf{Leveraging priors with QD state-of-the-art.} Same principle as in Fig. \ref{fig:exp1_results}.  All the tested QD variants outperform standard prior-based sampling schemes. The pressure for quality optimization makes QD\_contact variants behave similarly as QD\_approach, such that the algorithm generates solutions that verify the usually hard-coded approach criterion (Fig. \ref{fig:exp2_nu_results}).}
  \label{fig:exp2_results}
\end{figure}


\begin{figure}[t]
  \centering
  \includegraphics[width=\columnwidth]{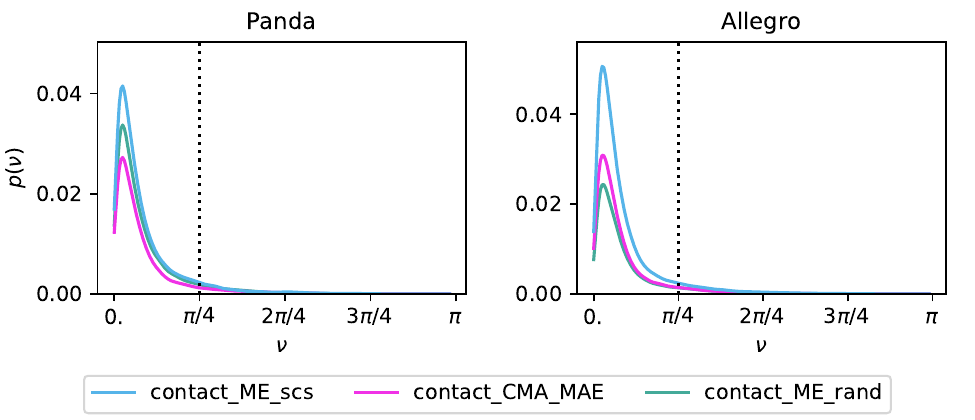}
  \caption{\textbf{Distribution of $\bm{\nu}$ from successful grasps generated with contact-based variants.} Each bin describes the probability \textit{p}$(\nu)$ of finding grasps poses with a $\nu$ angle between the normal to the hand palm and the normal at the nearest contact point on the object surface. By focusing on the most promising part of the search space, the QD method generates grasps that match the approach prior (i.e. $\nu\in [0, \pi/4]$).}
  \label{fig:exp2_nu_results}
\end{figure}


\begin{figure}[t]
  \centering
\centering
  \includegraphics[width=\columnwidth]{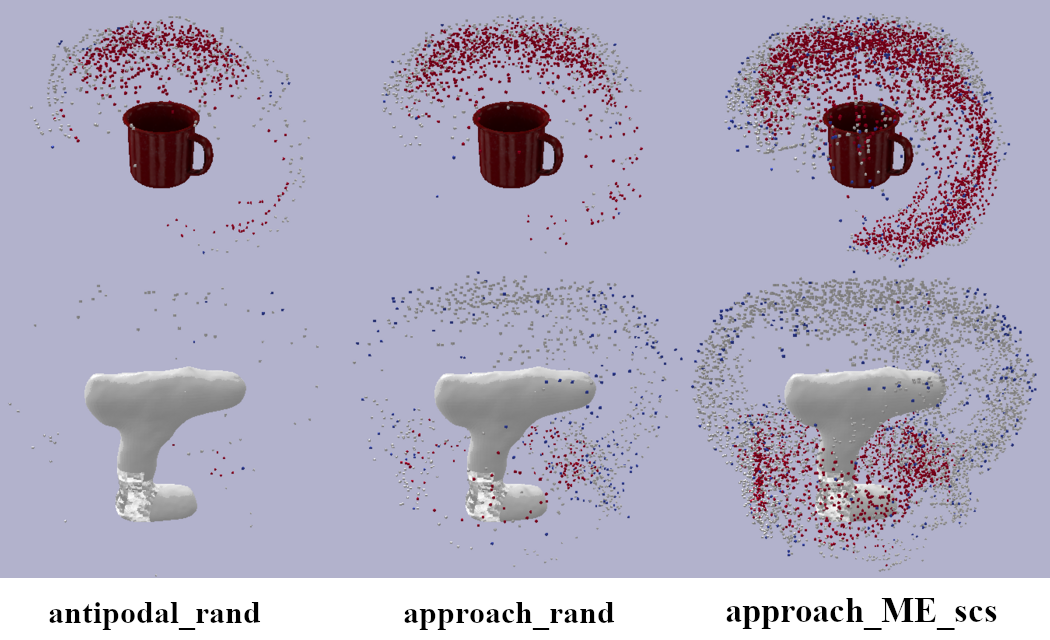}
  \caption{\textbf{Examples of successful grasps coverage per methods.} Each voxel is the end effector position of a simulated successful grasp. The heatmaps describe the robustness of each grasp – the hottest the voxel, the more robust the grasp. Leveraging QD results in a more sample-efficient search, eventually producing a set of grasp poses that cover the whole object surface.}
  \label{fig:visual_comparison_qd_with_without_priors}
\end{figure}


\textbf{\textit{Simulated Grasps Synthesis.}} Fig. \ref{fig:exp1_results} shows the evolution of the grasp space coverage with respect to the number of 6-DoF grasps samples evaluated in the simulation. The QD state-of-the-art method with limited priors (\textit{contact\_ME\_scs}) significantly outperforms the standard grasps sampling schemes (\textit{contact\_rand}, \textit{antipodal\_rand}, \textit{approach\_rand}) regardless of the platform. It shows that QD can speed up grasp sampling compared to standard methods. The usage of the contact prior is here critical, as it leads to a significant increase of performance compared to \textit{ME\_scs}. 

Interestingly, the results obtained on the standard schemes for the Panda gripper are lower than the one reported by Eppner et al. \cite{eppner2023abw2g}. This difference might come from the model of contacts in bullets and the simulation parameters that could make the grasps harder in Bullet than in FleX. Another difference is the grasp coverage metrics: Eppner et al. include the orientation of the gripper pose for measuring diversity, while we only keep the Cartesian position. This might shift the plateau of strongly constrained methods like \textit{antipodal\_rand} to a higher grasping coverage. However, the relative performances are comparable. 

Fig. \ref{fig:exp2_results} shows the evolution of the grasp space coverage for several combinations of state-of-the-art QD algorithms and robotic priors on the Panda gripper and the Allegro hand. QD variants with priors reached higher coverage than standard prior-based methods. Variants based on \textit{ME\_scs} are the best-performing ones, extending Huber et al. \cite{huber2023quality} results on reach-and-grasps trajectories to 6-DoF poses. The antipodal-based variant of \textit{ME\_scs} quickly plateaus, showing that this prior is too conservative to explore $\mathcal{G}$. Moreover, it can be noticed that QD variants with either the approach or the contact priors do not reach significant differences in coverage. This is unexpected, as there is a clear difference between standard random-based variants for approach and contact (see Fig. \ref{fig:exp1_results}). An explanation can be found in Fig. \ref{fig:exp2_nu_results}, which shows the distribution of $\nu$ variables for contact variants of QD methods. The distributions mostly lie between 0 and $\pi/4$, which is the constraint imposed by the approach prior. By focusing on the most promising part of the search space, QD variants generate grasp samples in which the palm normal is aligned with the normal on the object surface at the targeted contact point. In other words, QD contact-based variants automatically find priors that are usually hard-coded to speed up the generation of the grasps.

Fig. \ref{fig:visual_comparison_qd_with_without_priors} provides examples of FR3 grasps poses for standard 6-DoF grasp sampling scheme, as well as one of the two best-performing QD method, after 100k evaluation samples. The QD-generated grasps successfully cover the whole object's surfaces, including both fragile and robust grasp regions. It is well-known in the machine learning community that high-quality datasets include good and bad examples and do not ignore some parts of the targeted distribution. Several works build diverse grasp poses datasets using standard sampling scheme \cite{eppner2021acronym,eppner2023abw2g}. The obtained results suggest that similar datasets can be built significantly faster by leveraging QD-based optimization methods.

\begin{figure}[t]
  \centering
  \includegraphics[width=\columnwidth]{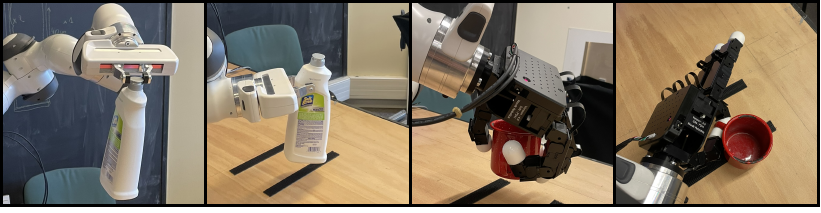}
  \caption{\textbf{Examples of transferred grasps.} The proposed method generates diverse grasp poses that are robust in the physical world.}
  \label{fig:real_grasps}
\end{figure}

\begin{table}[t]
\centering
\begin{tabular}{ ||c || c | c | c  ||}
\hline
QD Framework & Reach-and-grasp & \multicolumn{2}{c||}{6DoF}   \\
\hline
Gripper & Panda & Panda & Allegro \\
\hline
 $\eta^{sim2real}$  & 0.84 & \textbf{0.95} & \textbf{0.72} \\
\hline

\end{tabular}
\caption{\textbf{Measured sim2real transfer ratios.}}
\label{table:s2r_deploy_results}
\end{table}


\textbf{\textit{Real robot experiments.}} To evaluate the exploitability of the grasps generated by \textit{ME\_scs}, grasping repertoires have been generated with the same hyperparameters, except from the usage of a Domain-Randomization-based fitness that showed a significant alignment with sim2real transfer \cite{huber2023domainrandomization}. More details on the experimental setups can be found in Appendix \ref{sec:a2_exp_details_real}. Table \ref{table:s2r_deploy_results} reports the obtained sim2real transfer ratios $\eta^{sim2real}$ on a couple of objects when deploying the grasps with higher fitness in $A_s$. The 6DoF framework leads to comparable results with the reach-and-grasp framework, reaching 95\% of transfer on the Panda gripper. Deployments on the Allegro hand resulted in a lower ratio ($\eta^{sim2real}=72\%$). Most of the failures are due to 2-finger grasp synergies, which do not apply enough force for lifting the power drill. All the attempts succeeded on the bleach cleanser, the bowl, and the mug. In Fig. \ref{fig:real_grasps} are provided some examples of successfully transferred grasps. Those experiments show that the generated grasps can successfully be exploited in the physical world.


\textbf{Limitations.} The grasp generation experiments have been conducted with CPU parallelization only. Related works leverage GPU parallelization for scaling the generation of diverse 6-DoF grasp poses \cite{eppner2023abw2g}. Applying QD methods on GPU environments would imply a stochasticity inherent to parallelization. Recent works explore how to apply QD to stochastic environments \cite{chatzilygeroudis2021qdstochastic}. Exploring how to extend the present work to GPU parallelization is a promising way of improving the generation of diverse grasping datasets.

Another limitation is the grippers' control. There is no fine control of the force intensity here, and the dexterous hands are limited to synergy-based grasps. The use of simple sampling methods for dexterous hands is also restrictive. The proposed framework aims to speed up the generation of diverse and robust grasps for different grippers with minor adaptations. Exploiting multi-step optimization methods \cite{turpin2023fastgraspd,yao2023allegrograsp} within this framework might be a promising way to generate large datasets of diverse grasps for dexterous hands beyond synergies.


\section{CONCLUSIONS}

This work demonstrates the potential of combining Quality-Diversity methods with priors to speed up the 6-DoF grasp sampling schemes and build large datasets of diverse and robust grasps. Our proposed method can easily be applied to several robotic platforms, from parallel 2-finger grippers to dexterous hands. Experiments conducted on real robots show that the diversity of found solutions maintains sim-to-real transferability. We believe such a method can help the latest robotic learning models to converge to generalizing grasping policies by leveraging large datasets of simulated grasps.


\section*{ACKNOWLEDGMENT}

This work was supported by the Sorbonne Center for Artificial Intelligence, the German Ministry of Education and Research (BMBF) (01IS21080), the French Agence Nationale de la Recherche (ANR) (ANR-21-FAI1-0004) (Learn2Grasp), the European Commission's Horizon Europe Framework Programme under grant No 101070381 and from the European Union's Horizon Europe Framework Programme under grant agreement No 101070596. This work used HPC resources from GENCI-IDRIS (Grant 20XX-AD011014320). Many thanks to Louis Annabi and Olivier Serris for our insightful discussions.



\bibliographystyle{IEEEtran}


\clearpage
\appendices

\begin{center}
\textbf{\large Supplementary Materials}
\end{center}


\section{Experimental details: simulation}
\label{sec:a1_exp_details_sim}

\textbf{Objects.} Following YCB objects were used in the data generation experiments. For the Panda gripper: banana, bleach cleanser, bowl, cracker box, mug, power drill, rubiks cube, spatula, tennis ball; for the Barrett hand: banana, bleach cleanser, bowl, cracker box, power drill, rubiks cube, spatula, tennis ball; for the Allegro hand: banana, bleach cleanser, bowl, chips can, cracker box, mug, power drill, rubiks cube, spatula, tennis ball; for the Shadow hand: banana, bleach cleanser, bowl, mug, power drill, rubiks cube, spatula, tennis ball.

\textbf{Algorithms.} Let $\mu$ be the population size, $\lambda$ the number of offspring, $k$ the number of neighbors considered for novelty computation, and $N_e$ the maximum number of evaluation. We set: $\mu=\lambda=500$, $k=15$, $N_e=100\text{k}$. All offspring are mutated with a probability $ind_{pb}=0.3$ to modify each gene. For a fair comparison, all ME-derivated methods sample $\mu=\lambda$ individuals for offspring generation at each iteration. The mutation operator applied by default to all the methods is a Gaussian perturbation of $0$ mean and $0.1$ standard deviation. This value of $\sigma$ has been obtained by doing grid search, trying to get the highest possible value of $cvg(\mathcal{G})$ after 20k evaluated samples. The variance of optimal value of $\sigma$ between methods was so small that we set it to 0.1 for all the mutation-based methods. For \textit{CMA\_MAE} variants, we used the same parameters as in \cite{fontaine2023cmamae}: The emitter batch size is set to $36$, and the number of emitters to $15$, $f_{min} = -1$ and $\alpha=0.01$.

The fitness $f$ is computed by applying a predefined shaking pattern, similarly to \cite{eppner2023abw2g}. The fitness of a grasp pose $g$ is equal to the number of shakes applied to the gripper after the closure without making the object fall. Let $N_s$ the number of performed shakes. We set $N_s=2$: the first perturbation is a translation shake, the second one is a rotation - both around a fixed axis in the simulated world. Therefore, the optimal fitness is $f^*=N_s=2$.

\textbf{Grippers.} Following are the details on the way each gripper is controlled. For the Panda gripper: The fingers opened as wide as possible, and closed at grip time; for the Barrett hand: the initial position of the joints that orient the proximal part of each fingers with respect to the palm are defined by the parameter, such that the search space allows to get several initial configuration of those fingers. All the fingers are then uniformly closed; for both the Allegro and the Shadow hand: the hand is closed with respect to synergies. The synergies consist of a uniform closure of specific fingers, including thumb-index, thumb-mid, thumb-index-mid, and all the fingers. All fingers are initialized in wide open position. The thumb palm-to-proximal joint is always set such that the thumb is orthogonal to the palm. This kind of synergies have been used for their simplicity, their efficiency, and the ease with which they can be adapted to different grippers. Other kind of synergies can be found in the literature, like the eigen grasps \cite{miller2004graspit}. Studying the optimal way to generate grasps for n-fingers is beyond the scope of this paper.


\section{Experimental details: physical world}
\label{sec:a2_exp_details_real}

\textbf{Data generation.} The fitness has been replaced with a quality criterion dedicated to sim2real transfer. We use a variant of the Mixture Domain Randomization criterion proposed in \cite{huber2023domainrandomization}, removing the variance on the joint state - as it is redundant with the variance on the object state in the 6DoF grasp prediction context. Let $N_{MDR}$ be the number of retrials, $\sigma_{o_p}$ the object position variance along each axis, and $\sigma_{o_o}$ its orientation variance. We set $N_{MDR}=100$, $\sigma_{o_p}=0.005 \text{m}$, $\sigma_{o_o}=30^{\circ}$. We kept the same variance of friction coefficient as in \cite{huber2023domainrandomization}. The shaking process is similar to the simulated experiments. Consequently, the best possible fitness is $f_{MDR}^*=N_s \times N_{MDR} = 200$. The probability $\eta^{sim2real}$ for a given grasp $g$ to successfully transfer in the physical world can thus be estimated as:
\begin{equation*}
    \eta^{sim2real}(g)=\frac{f(g)}{f_{MDR}^*}
\end{equation*}

All the remaining hyperparameters are similar to those used in the simulated experiments. The grasps are then sorted with respect to their fitness $f_{MDR}$. 

\textbf{Real world setup.} Real experiments involve a Franka Research 3 arm with a parallel gripper and an Allegro-Hand V4 with gravity compensation and some compliance. The same arm is used for both grippers. A 3D-printed mounting adaptor is used to fix the Allegro hand on the arm. The Allegro hand is controlled with a joint impedance controller with gravity compensation. No additional material that could increase adhesion is used to make grasping easier. The synergies are the same as the ones used in the simulated experiments. The gravity compensation is computed with respect to the fixed arm base. The motion planning of the end effector is conducted with RRT connect through \textit{Moveit!}, assuming that the object and the table are collision bodies. The object pose detection is conducted by leveraging a hand-eye calibrated Realsense D435i, using the vision pipeline from Hélénon et al. \cite{helenon2023fsta}. The angle between the camera point-of-view and the table plan is roughly $30^\circ$.

\textbf{Experiments.} The real experiment involves 4 YCB objects: the bowl, the mug, the bleach cleanser and the power drill. It is worth noting that among the sorted high-performing grasps, the ones that do collide with the table were discarded. Almost all the deployed trajectories have successfully been transferred on the physical gripper. The failures are primarily due to a misalignment between the applied forces in simulation and the real world. This also raises a limitation of the 6DoF setup widely used in the literature. A finer control of applied forces could overcome this limitation.

\end{document}